%% file: root.tex
\documentclass[letterpaper, 10 pt, conference]{IEEEtran}  %

\IEEEoverridecommandlockouts                              %

\input{include/include_packages.tex}

\input{include/include_macros.tex}

\usepackage{svg}

\usepackage{mathptmx} %
\usepackage{subfigure}
\usepackage{xcolor}
\newcommand{\new}{{}}
\newcommand{\fix}{{}}

\input{include/alaa_macros}

\definecolor{flodarkpurple}{rgb}{0.288,0.1196,0.7}
\usepackage[backref=page,breaklinks,colorlinks,bookmarks,citecolor=flodarkpurple]{hyperref}

\newcommand{\FAM}{\textit{FAn}}

\begin{document}
\title{\textsc{\textit{Follow Anything}}: Open-set detection, tracking, and following in real-time%
}

\author{\href{https://alaamaalouf.github.io/}{\color{violet}Alaa Maalouf$^{1,2,3,*}$}, \href{https://react.seas.harvard.edu/people/ninad-jadhav}{\color{violet}Ninad Jadhav$^{2,3}$}, \href{https://krrish94.github.io/}{\color{violet}Krishna Murthy Jatavallabhula$^{1}$}, \href{https://www.mit.edu/~chahine/}{\color{violet}Makram Chahine$^{1}$},\\  \href{https://www.danielmvogt.com/}{\color{violet}Daniel  M.Vogt$^{2,3}$}, 
 \href{https://wyss.harvard.edu/team/associate-faculty/robert-wood/}{\color{violet}Robert J. Wood$^{2,3}$}, \href{https://groups.csail.mit.edu/vision/torralbalab/}{\color{violet}Antonio Torralba$^{1,3}$}, and \href{https://danielarus.csail.mit.edu/}{\color{violet}Daniela Rus}$^{1,3}$ \\ \\
 \href{https://www.csail.mit.edu/}{\color{magenta}$^1$CSAIL, MIT} \quad \href{https://seas.harvard.edu/}{\color{magenta}$^2$SEAS, Harvard University} \quad \href{https://www.projectceti.org/}{\color{magenta}$^3$Project CETI}\quad {\color{blue}$^*$Correspondence: alaam@mit.edu}}%

\makeatletter
\let\@oldmaketitle\@maketitle
\renewcommand{\@maketitle}{\@oldmaketitle
\centering
\includegraphics[width=0.98\linewidth]{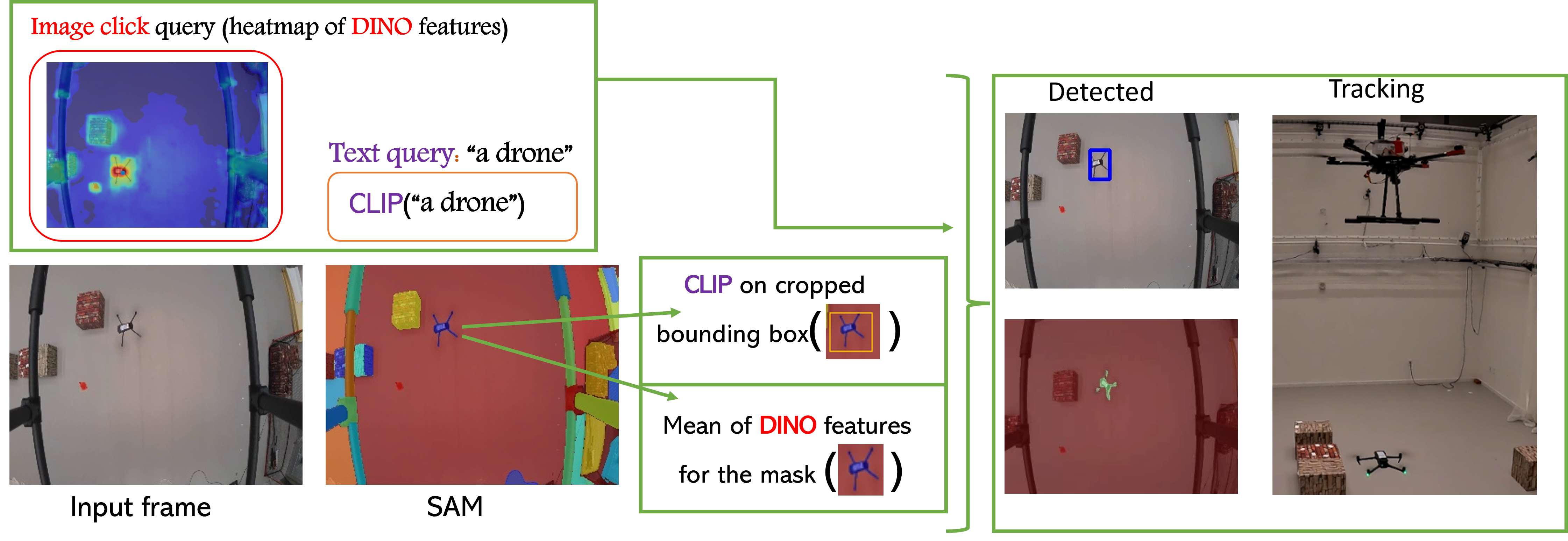}
\captionof{figure}{\textbf{Follow anything} (\FAM{}) is a real-time robotic system to detect, track, and follow objects in an open-vocabulary setting. Objects of interest may be specified using text descriptions, images, or clicks. \FAM{} then leverages foundation models like CLIP~\cite{radford2021learning}, DINO~\cite{caron2021emerging}, and SAM~\cite{kirillov2023segment} to compute segmentation masks and bounding boxes that best align with the queried objects. These objects are tracked across video frames, accounting for occlusion and object re-emergence; enabling real-time following of objects of interest by a robot platform. On the right, we show \FAM{} being deployed on a micro aerial vehicle (MAV), following an object of interest (here, another smaller \emph{drone}). \FAM{} can be deployed on a commodity laptop with a lightweight (6-8 GB) GPU, achieving real-time (6-20 fps) throughput. \new{We encourage the reader to check our \href{https://www.youtube.com/watch?v=6Mgt3EPytrw}{explainer video} and to view the demos on our project webpage: \href{https://github.com/alaamaalouf/FollowAnything}{https://github.com/alaamaalouf/FollowAnything}}.
}
\label{fig:all}
}

\maketitle

\begin{abstract}
Tracking and following objects of interest is critical to several robotics use cases, ranging from industrial automation to logistics and warehousing, to healthcare and security. 
In this paper, we present a robotic system to detect, track, and follow any object in real-time. 
Our approach, dubbed \emph{follow anything} (\FAM{}), is an open-vocabulary and multimodal model -- it is not restricted to concepts seen at training time and can be applied to novel classes at inference time using text, images, or click queries.
Leveraging rich visual descriptors from large-scale pre-trained models (\emph{foundation models}), \FAM{} can detect and segment objects by matching multimodal queries (text, images, clicks) against an input image sequence.
These detected and segmented objects are tracked across image frames, all while accounting for occlusion and object re-emergence.
We demonstrate \FAM{} on a real-world robotic system (a micro aerial vehicle), and report its ability to seamlessly follow the objects of interest in a real-time control loop.
\FAM{} can be deployed on a laptop with a lightweight (6-8 GB) graphics card, achieving a throughput of 6-20 frames per second.
\new{To enable rapid adoption, deployment, and extensibility, we open-source our code on our \href{https://github.com/alaamaalouf/FollowAnything}{project webpage}. We also encourage the reader to watch our 5-minute \href{https://www.youtube.com/watch?v=6Mgt3EPytrw}{explainer video}}.
\end{abstract}

\input{text/01-introduction}

\input{text/02-approach}

\input{text/03-experiments}

\input{text/04-discussion-and-conclusion}

\bibliographystyle{IEEEtran}
\bibliography{IEEEtranBST/IEEEabrv, root}

\end{document}

%% file: include/include_packages.tex
\usepackage{adjustbox}
\usepackage[ruled, vlined]{algorithm2e}
\usepackage{amsmath}
\usepackage{amssymb}
\usepackage{array}  %
\usepackage{authblk}
\usepackage{booktabs}
\usepackage{capt-of}
\usepackage{cite}
\makeatletter
\let\NAT@parse\undefined
\makeatother
\usepackage{cite}
\usepackage[font={small}]{caption}
\usepackage{colortbl}
\usepackage{epsfig}
\usepackage{inconsolata}  %
\usepackage{graphicx}
\usepackage{listings}
\usepackage{multicol}
\usepackage{multirow}
\usepackage{paralist}
\usepackage{pifont}
\usepackage{times}
\usepackage{tabularx}
\usepackage{xcolor}
\usepackage{svg}

%% file: include/include_macros.tex
\definecolor{ganeshamber}{rgb}{1.0, 0.75, 0.0}

%% file: include/alaa_macros.tex
\newcommand{\seg}{\texttt{Seg}}
\newcommand{\desc}{\texttt{Desc}}
\newcommand{\nnz}{\texttt{non-zero}}
\newcommand{\norm}[1]{\left\lVert#1\right\rVert}%

\newcommand{\eps}{\varepsilon}

\newcommand{\br}[1]{\left\lbrace #1 \right\rbrace}

\newcommand{\abs}[1]{\left| #1 \right|}
\newcommand{\REAL}{\mathbb{R}}

\newcommand{\M}{M^{obj}}

%% file: text/01-introduction.tex
\begin{figure*}[t]
    \centering
    \includegraphics[width=0.98\linewidth]{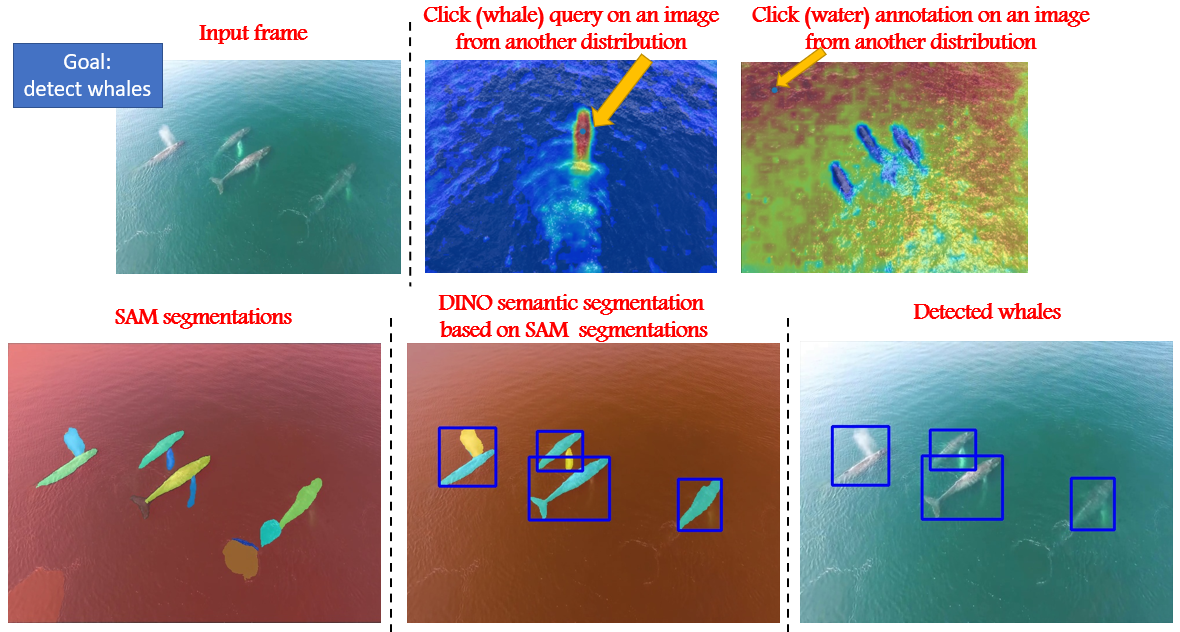}
    \caption{\FAM{} outputs illustrations on input frame of $4$ whales with a click query on a whale and a click query on water. First, SAM extracts multiple masks, then, based on DINO features, \FAM{} classifies each mask to what object it refers to from the given queries (water/whales). Finally, whales are detected by assigning the masks whose DINO feature descriptor is closest to the whales' query descriptor. NOTE: Heat maps are shown in the click (query) figures.}
    \label{fig:example}
\end{figure*}
\section{Introduction}
Detecting, tracking, and following objects of interest is critical to several robotics use-cases, such as industrial automation, logistics and warehousing, healthcare, and security~\cite{8846214,maalouf2022deep,naeem2013real,koubaa2018dronetrack,chen2019real}. Notably, one of the key drivers of continuous progress in providing robust object-following systems is the combination of computer vision and deep learning~\cite{restas2015drone,tezza2019state,maalouf2022deep}, where training deep convolutional networks on large labeled datasets have made tremendous strides in this area.  

\fix{Specifically, the object following task relies on the video segmentation and tracking task, which can be categorized into distinct subtasks. These include interactive (scribble or click-based) video segmentation~\cite{cheng2021modular}, where a user draws a box around or clicks on the object to segment and track, mask-guided video segmentation~\cite{lu2020video,yang2022scalable,yang2021associating,yang2022decoupling}, which assumes the presence of a mask to track, and automatic video segmentation~\cite{cho2023treating,lu2019see,wang2019zero,wang2019learning,cheng2023segment}, which assumes that the user does not interact with the algorithm to obtain the segmentation masks; methods should provide a set of object candidates with no overlapping pixels that span through the video sequence, however, these candidates are not specific, meaning that the segmentation will be applied to all of the seen objects, and not recognize the desired object. Thus, to automatically identify the required object to follow, numerous detection approaches have been suggested~\cite{dai2016r,tan2020efficientdet} such as RCNN and its variants~\cite{girshick2014rich,girshick2015fast,ren2015faster}, YOLO and its variants~\cite{redmon2016you,bochkovskiy2020yolov4,redmon2018yolov3}, and more~\cite{liu2016ssd,xing2023go}.} However, existing robotic systems for object detection and following suffer two notable shortcomings: 
    (i) They are \emph{closed-set}, i.e., the set of objects to detect and follow is assumed to be available a priori (during the training phase).  %
    This means such systems are only able to handle a \emph{fixed set of object categories}~\cite{maalouf2022deep,9043931,9785379,ganti2016implementation,unlu2019deep}, limiting their adaptability; adapting to newer object categories necessitates finetuning the model. 
    (ii) Additionally, the objects of interest are specified (queried) only by a class label, which is \emph{often unintuitive for end-users} to specify, imposing restrictions on how users interact with the system~\cite{7429411,maalouf2022deep,8999675}.

Deep learning is currently undergoing another wave of ever-more performant and robust model design, with the creation of increasingly big and multimodal models trained on internet scale amount data containing billions of images, text, and audio. These highly capable models  (e.g., CLIP \cite{radford2021learning}, DINO \cite{caron2021emerging}) have demonstrated impressive performance in open-set scenarios (i.e., the objects of interest are only supplied at inference time, and not trained for a specific task)~\cite{jia2021scaling,guzhov2022audioclip}. Notably, recent robotics approaches using foundation models have shown impressive open-set interaction abilities~\cite{tellex2020robots,bisk2020experience,ahn2022can,brohan2022rt,li2022pre,wang2023drive}, and extended robustly to multimodal applications~\cite{ramesh2021zero,wang2023drive,crowson2022vqgan,patashnik2021styleclip,ramesh2022hierarchical}. 
 However, integrating these performant models into real-time and resource-constrained robotic systems poses significant challenges, due to their large model size and high inference latency.

\subsection{Our {Contributions}}
We address the pre-discussed gaps by developing an \emph{open-set} real-time any object following approach, which can flexibly adapt to categories specified at inference time, via multiple modalities including text, images, and clicks. Specifically, we present the \emph{follow anything} system (\textbf{\FAM}):
\begin{itemize}
    \item an \textbf{open-set}, \textbf{multimodal} approach to detect, segment, track, and follow \emph{any} object in \textbf{real-time} ($>6$FPS on a 8GB GPU).
    The desired object may be specified via a text prompt, an image, a bounding box, or a click.
    \item a \emph{unified} system that is easily deployed on a robot platform (in our work, a micro aerial vehicle). The system includes real-time processors for input image streams and visual-servoing control loops for following the object of interest.
    \item built with \emph{re-detection mechanisms} that account for scenarios where the object of interest is occluded or tracking is lost. This mechanism can function autonomously or with human guidance, ensuring the object is successfully identified and tracked again, maintaining continuity in the tracking process.
\end{itemize}

We validate our system by autonomously detecting, tracking, and following a multitude of mobile agents including a drone, an RC car, and a manually operated brick.

%% file: text/02-approach.tex
\section{Our Approach: \FAM{}}

\noindent\textbf{Open-vocabulary object following}:
Given (1) a robotic system (here, a micro aerial vehicle) equipped with an onboard camera, and (2) an object of interest within the onboard camera's field-of-view (specified either as a text prompt, an image, a bounding box, or a click); the object following task involves detecting the object of interest, and producing robot controls $u_t$ at each time step $t$ such that the object of interest is constrained to always completely be within the field of view of the onboard camera. 
This is an extremely challenging task; it necessitates correctly identifying the object of interest and determining its position relative to the robot's onboard camera frame, all the while accounting for variations in the environment, background clutter, object size, etc. It also then requires the object to be continuously tracked across time; while at the same time, the robot controller needs to output a sequence of stable velocities (or accelerations) and simultaneously ensure the stability of the robot and the visibility of the tracked object.

\noindent\textbf{\FAM{} system overview}:
\FAM{} uses a combination of state-of-the-art ViT models, optimizes them for real-time performance, and unifies them into a single system. In particular, we leverage the segment anything model (SAM)~\cite{kirillov2023segment} for segmentation, DINO~\cite{caron2021emerging}, and CLIP~\cite{radford2021learning} for general-purpose visual features, and design a lightweight detection and semantic segmentation scheme by combining the features from CLIP and DINO with the class-agnostic instance segmentation determined by SAM. 
We use the (Seg)AOT~\cite{yang2022deaot,yang2021aost} and SiamMask~\cite{wang2019fast} models for real-time tracking, and design a lightweight visual servoing controller for object following.

\subsection{Real-time open-vocabulary object detection}
\begin{figure*}[!t]
    \centering
    \includegraphics[width=1\linewidth,height=0.5\linewidth]{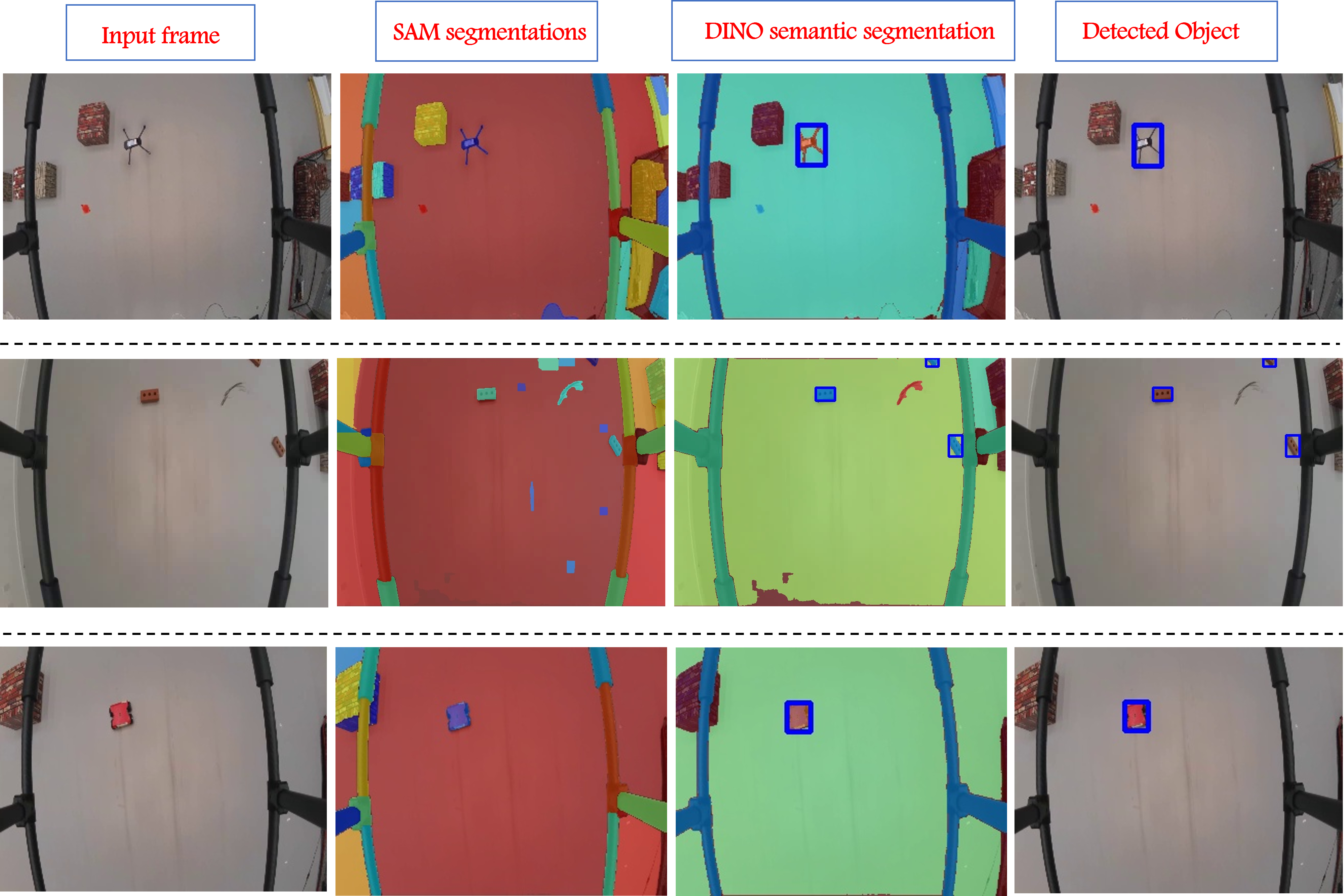}
    \caption{\textbf{Automatic detection experiments (SAM-and-DINO).} Examples of our automatic detection scheme for detecting Drones, Bricks, and RC Cars. The examples include (from left to right): the original input frame, the outputs of SAM segmentation masks, and DINO+Cosine similarity semantic segmentation and detection.}
    \label{fig:autodetect}
\end{figure*}

We first describe our lightweight object detection and segmentation pipeline that builds atop SAM, CLIP, and DINO.
Our system takes as input an RGB frame from a video stream, represented by a 3D tensor $F \in \REAL^{h\times w \times 3}$, and a query $q$ representing the desired object to detect in the video, (e.g., a text ``\emph{a blue whale}'', an image of a whale, or a click on a whale from another image). 
The object detection subsystem is tasked to detect the object specified by the user query $q$  in an input image frame. 
We use \seg{} to denote the class-agnostic instance segmentation operator (SAM~\cite{kirillov2023segment} or Mask2Former~\cite{cheng2022masked}).
\seg{} takes as an input the current frame $F$, and outputs a set of $n$ masks $\br{M_1, \cdots, M_n}:= \seg(F)$ ($n$ depends on the input frame and is not a constant), where each mask $M_i \in \REAL^{h\times w}$ is a binary matrix with ones in the indices of pixels defining the corresponding segmented object, and zeros elsewhere.
We also use \desc{} to denote a feature extractor model; which in our case is either the DINO or CLIP vision transformer (ViT) model. These models extract pixel-wise feature descriptors using techniques described e.g., in~\cite{amir2021deep,jatavallabhula2023conceptfusion} and summarized in Section~\ref{sec:pixel-wise feature}.
\desc{} receives as an input the current frame $F$, and outputs a descriptor tensor $D:=\desc(F)\in \REAL^{h\times w \times d}$, where for every pixel in $F$, a descriptor vector of dimension $d$ is constructed. This  $d$ dimensional vector encapsulates the semantic information about its corresponding pixel. Additionally, \desc{} can be also used to provide a feature descriptor $v := \desc(q)\in\REAL^{d}$ for the input query $q$.

\noindent\textbf{Embedding input queries.} To detect the desired object referred to by the query $q$, we start by computing the feature descriptor of the query $q$:
$v := \desc(q),$
such that $v$ encodes the information in the feature space representing the object described by the query $q$. Now the system starts receiving frames from the stream, and for every frame $F_i$ ($i:= 1,2,3\cdots$), \FAM{} applies the following steps. 

First, we compute the (binary) instance segmentation masks by applying $\seg$ on $F_i$,
$\br{M_1, \cdots, M_n}_i:= \seg(F_i)$.
Intuitively, this step partitions the frame into $n$ objects (regions) and a background, however, none of these objects are classified as labeled/identified objects. Additionally, these regions might intersect. 
Hence, what is missing, is to predict for each region, whether it is the desired object to track or not. In the case where a set of queries $Q = \br{q_1, \cdots, q_m}$ is given, the goal is to classify which query amongst the $m$ provided matches (if any) best each segment $M_j\in  \seg(F_i)$. This brings us to the second step.

Second, we extract the pixel-wise descriptors by applying $\desc$ on $F_i$,
$D_i:=\desc(F_i)\in \REAL^{h\times w \times d}$.
After this step, $D_i$ contains $h\cdot w$ descriptors, where each descriptor corresponds to a pixel in the input image. To compare each region (mask) with the given input query $q$, we need to aggregate these per-pixel descriptors to form region-level descriptors.
We find the average pooling aggregation operator to be fast and effective for this purpose.
This not only provides us with a more generic descriptor encapsulating all of the features across the specific mask but also improves the performance of the downstream system modules. Opposed to comparing the $q$ query's feature descriptor $v$ to all of the per-pixel descriptors associated with a specific mask, we only need to compare the aggregate region-level descriptors.
Thus, the next step in our pipeline involves computing the mean feature descriptor $v_j$ for each segmentation region, i.e., for every $j\in \br{1,\cdots,n}$: 
$$v_j:= \frac{1}{\nnz({M_j})}\sum_{p\in D_i[M_j]} p, $$
where $\nnz(M)$ denotes the number of non-zero entries in binary matrix $M$, and $D_i[M_j]$ denotes the set of $d$ dimensional vectors from $D_i$ corresponding to the non-zero pixel entries in the mask $M_j$. The vector $v_j$, encodes the semantic information representing the region of the segment $M_j$ in the features space. For every region (segment/mask) $j\in \br{1,\cdots,n}$, we have its corresponding descriptor $v_j$. 

\noindent\textbf{Similarity scores}:
Given a query (in the form of text, image, or click), we first extract a query feature descriptor $v$ by applying a modality-specific encoder (CLIP for text-query, DINO or CLIP image encoder followed by average pooling for image-query, directly selecting the closest pixel/patch feature for click-query).
To match this query to the current image, we compute the cosine similarity between each region descriptor $v_j$, and the query feature descriptor $v$ as 
$$\cos(v_j,v):=\frac{v_j^T v}{\norm{v_j}\norm{v} + \varepsilon},$$
where $\eps>0$ is a small constant, for numerical stability. This is the fourth step, and it intuitively measures how similar each mask (region) is to the query features descriptor.

\noindent\textbf{Single query detection.} If the similarity  $\cos(v_j,v)$ between the given query and the mask feature descriptor is larger than a given threshold $\alpha$, we assign the region corresponding to this mask in the original frame the label of the query. 

\noindent\textbf{Multi-class detection.}  %
Should the user provide a set of queries $Q = \br{q_1, \cdots, q_m}$, the system computes the descriptor $v^k:= \desc(q_k)$ for every $q_k\in Q$, then, for every pair of query descriptor $v^k$ and region descriptor  $v_j$, it computes: $\cos(v_j,v^k)$. Now, for every $j\in \br{1,\cdots,n}$, it finds its most similar query:
$$\max_{k\in \br{1,\cdots,m}} \cos(v_j,v^k).$$
Finally, if the cosine similarity between the query vector ($v^k$), and the mask descriptor ($v_j$), exceeds a threshold $\alpha$, we assign the label of the query to the region in the original frame corresponding to this mask, otherwise, it is considered "non-labeled".

After this process, each pixel is assigned a label from $\br{1,\cdots,m}$, or $0$ if unlabeled. Figure~\ref{fig:example} provides an illustration of the whole detection flow, and Figures~\ref{fig:autodetect} and \ref{fig:clipdetect} present results on detecting objects via SAM+DINO, and SAM+CLIP respectively.

\noindent\textbf{Manual queries}: We provide the users an option to manually draw bounding boxes (or provide outputs from a customized domain-specific detector) around the objects they wish to track, or alternatively, click on one or two pixels within the object (in real-time from the video stream). After user selection, we use SAM to accurately segment and obtain the object mask. This method ensures precise control over tracking, making it suitable for high-accuracy detection scenarios.
\begin{figure*}[!t]
    \centering
    \includegraphics[width=1\linewidth]{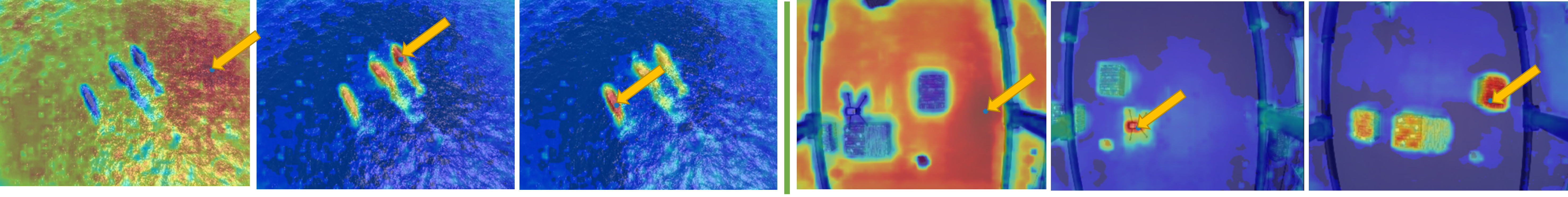}
    \caption{Heat maps showing the pixels' semantic similarity. For every pixel, its feature descriptor is extracted then cosine similarity is computed between its descriptor and a focal point pixel descriptor (pointed at by a yellow arrow).}    \label{fig:heatmaps}
\end{figure*}

\subsection{Fast detection for limited hardware}

Off-the-shelf implementations of foundation models like SAM and DINO are not well-suited for real-time onboard detection, segmentation, and tracking.
SAM takes several seconds to compute segmentation masks per frame. While we evaluated the recently proposed FastSAM~\cite{zhao2023fast} model and obtained a $15\times$ speedup on our hardware with comparable performance, the best runtime achieved by FastSAM is between 10 and 12 FPS, which is still insufficient for detecting fast-moving objects.
This is because segmentation outputs also need to be supplemented by features from ViT models, and the detection submodules.

\noindent\textbf{Fast detection by (solely) grouping DINO features}:
To mitigate this compute bottleneck, we instead propose to first obtain coarse detections by grouping DINO features. These coarse detections may further be refined by periodically computing segmentation masks and tracking these over time, effectively rendering the overall system operable at high frame rates.
To obtain coarse detections, (i) we extract the pixel-wise descriptors by applying $\desc$ (DINO) on the current input frame $F_i$, $D_i:=\desc(F_i)\in \REAL^{h\times w \times d},$ (ii) given the inputs set of queries $Q = \{q_1, \cdots, q_m\}$, the system computes the cosine similarity $\cos(v_{h,w},v^k)$ for each pair of query $q_k \in Q$ (where $v^k:= \desc(q_k)$) and pixel-wise descriptor vector $v_{h,w}$. 

\begin{figure}
    \centering
    \includegraphics[height = 0.16\linewidth,width=1\linewidth]{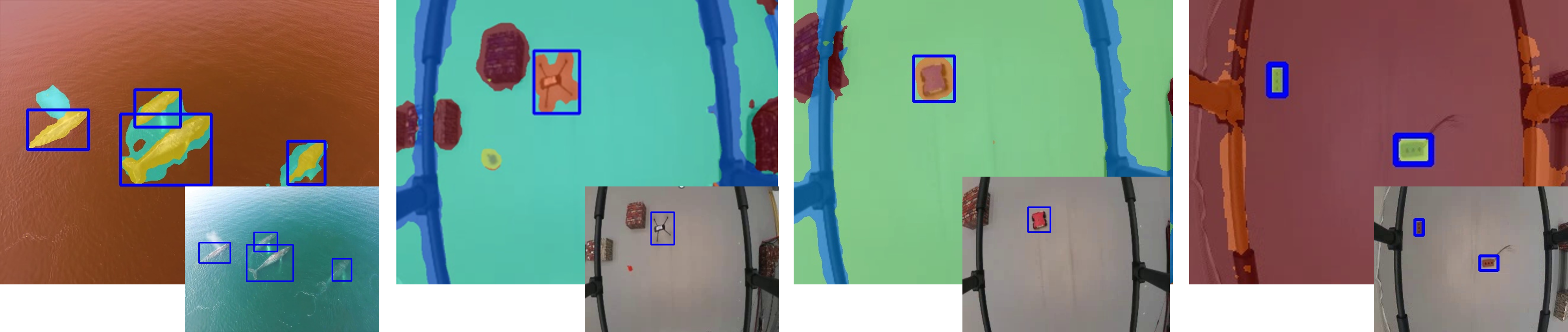}
    \caption{\textbf{Fast automatic detection experiments (DINO only):} Examples of our fast automatic detection scheme on detecting (1) whales, (2) drones, (3)  RC cars, and (4) toy bricks. This approach is much faster and works very well for detecting the desired object. However, it provides a less "clean" segmentations/masks.}
    \label{fig:fastapp}
\end{figure}

Next, as previously, (iii) for each pixel, it picks the closest (most similar) query, i.e., the one with the maximum cosine similarity.  
Now, (iv)  if the cosine similarity between the query vector $v^k$ and the pixel feature descriptor $v_{h,w}$ surpasses a specified threshold $\alpha$, we assign the label of the query to the corresponding pixel in the original frame $F_i$. Otherwise, it is considered as "non-labeled". \new{Then, (v) we build a binary matrix $B_i\in \REAL^{h\times w}$ with $1$ in pixels that are mapped to the desired object (to detect) and $0$ elsewhere. Finally (vi) apply the \texttt{cv2.connectedComponents} function on $B_i$. This function receives a binary image ($B_i$) where white regions (pixels with label 1) on a black background (pixels with label 0) represent connected components. The function assigns unique integer labels to each connected component and labels background pixels as 0. We have used it since we might detect more than one object, each in a different region of the frame, this function provides us with each object with its unique mask.} See Figure~\ref{fig:fastapp} for experiments leveraging the detection module proposed here.

\noindent\textbf{Optimizing DINO runtime}: We speed up DINO using two optimization techniques: Quantization (reduces numerical precision) and tracing (converts dynamic graphs into static ones). 
See Table~\ref{table:FPS} for runtime details of all the used models in our system. We report the running time for each model independently, not as part of the whole system. Note that some models automatically reshape inputs to a constant size. We also compare the runtime of our detection phase, with the popular Grounded-SAM~\cite{liu2023grounding} method in Table~\ref{table:compete}.  Further runtime improvements can still be made via automatic compression methods such as~\cite{liebenwein2021compressing,maalouf2022unified,liu2020autocompress}

\subsection{Extracting per-pixel feature descriptors}\label{sec:pixel-wise feature}

While a few methods adapt foundation models like CLIP to provide per-pixel descriptors,
these methods~\cite{li2022language,zhao2017open,ghiasi2022scaling,zhong2022regionclip} require model re-training or finetuning on an image-text aligned dataset.
This often results in concepts absent in the finetuning set being forgotten by the models as demonstrated in ConceptFusion~\cite{jatavallabhula2023conceptfusion}. %
 To counteract this, \cite{jatavallabhula2023conceptfusion} presents a zero-shot method for constructing pixel-aligned features that combine local (region-level) data with global (image-level) context included in models like CLIP.
For efficiency (real-time processing) purposes, we adapt part of this method in our system when using CLIP for providing pixel-wise feature descriptors, however, we only use their ablated baseline which computes purely local 2D features by extracting a bounding box around each segmentation mask (obtained from SAM) and passes them through the CLIP encoder.
For DINO, we use~\cite{amir2021deep} as is, and find that their pixel-wise feature descriptors are inherently informative and more efficient.%

\subsection{Re-detecting a lost object}
\begin{figure}[t]
    \centering
    \includegraphics[width=1\linewidth]{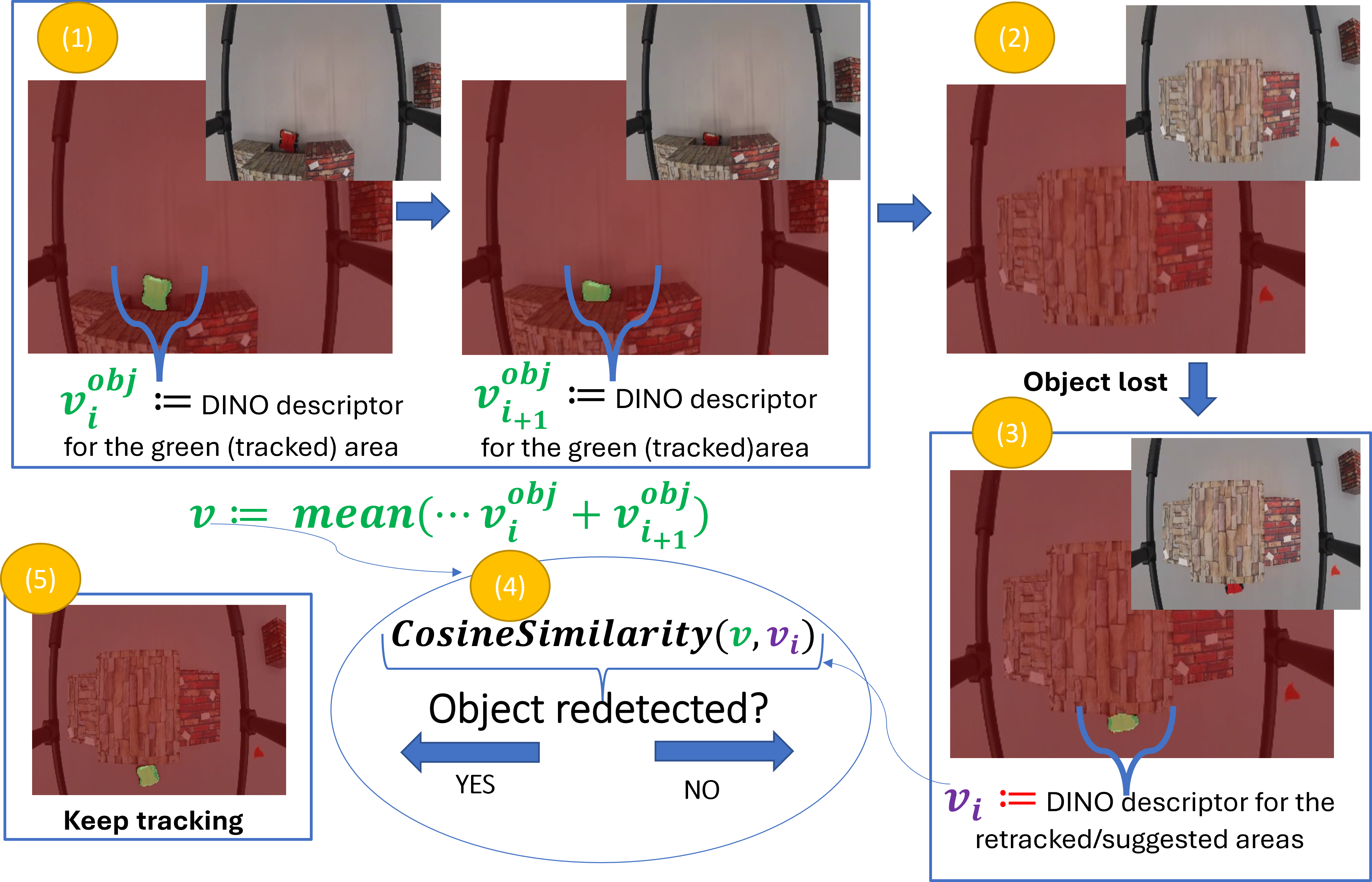}
    \caption{\textbf{Automatic re-detection via cross trajectory stored ViT features.} (1) At every frame, we store the DINO features representing the tracked object. (2) Once the object is lost, we  (3) either apply a segmentation model or get suggested masks from the tracker, for every mask, we compute the DINO descriptors, and (4) compare it to the pre-computed ones. If a high similarity is obtained we keep tracking the object, else, we repeat (3) on the next frame.}
    \label{fig:redetect}
\end{figure}

We offer three re-detection methods for temporary object loss during tracking, catering to different needs. Our system automatically initiates re-detection when needed, and users can choose the level of support before starting the \FAM{} pipeline: The first level relies on the tracker to re-detect the object, it's the fastest and less robust, occasionally leading to false detections of similar objects. The second approach involves human-in-the-loop re-detection, requiring a user to click/draw a bounding box when tracking is lost, assuming human availability, which isn't always possible. To mitigate this, we also propose an automatic re-detection technique.

\noindent\textbf{Automatic re-detection via cross trajectory stored ViT features.} 
To enable a robust and accurate autonomous re-detection of the tracked (lost) object, we provide a feature-descriptor storing mechanism for the tracked object in different stages of the tracking process, these stored features, will be used to find the object once lost. Specifically, we suggest the following. Let $\tau > 0$ be an integer. During the tracking, at each iteration $i$ such that $i\mod\tau = 0$, define $\M_{i}$ to be the mask denoting the current tracked object in the frame, we first apply \desc{} on the current frame $F_i$ to obtain $D_i:=\desc(F_i)\in \REAL^{h\times w \times d}$, then we compute the mean descriptor of the current tracked object as: 
$$v^{obj}_i:= \frac{1}{\nnz(\M_i)}\sum_{p\in D_i[{\M_i}]} p.$$ This feature represents the tracked object in the $i$th step. We thus store this descriptor and add it to the set of previously computed descriptors to obtain the set 
$$V^{obj}:= \br{v^{obj}_0, v^{obj}_\tau,v^{obj}_{2\tau},\cdots,v^{obj}_{i}}.$$ 
Now, whenever the system loses the tracked object, we apply the following recovery mechanism. The system goes back to the detection stage, with a query feature descriptor at hand as 
$$v = \frac{1}{\abs{V^{obj}}}\sum_{v^{obj} \in V^{obj}} v^{obj},$$
seeking the closest region from the segmented frame, and thus re-detecting the object. Here, the segmentation might be given by the segmentation model (e.g., SAM), or by the tracker which tries to re-detect the lost object.
Note we use the mean to gain faster performance for real-time applications, however, other techniques can be used to improve the robustness; see Figure~\ref{fig:redetect}.

\begin{table}[t]
\centering
\caption{Runtime in frames per second (FPS) for all of the used models on an NVIDIA GeForce RTX 2070 onboard a laptop.}\label{table:FPS}
\adjustbox{max width=0.5\textwidth}
{
\begin{tabular}{lccl}
\toprule
\multicolumn{1}{c}{} & \multicolumn{1}{c}{\textbf{FPS}} & \multicolumn{1}{c}{\textbf{FPS}} & \multicolumn{1}{c}{} \\  
\multicolumn{1}{c}{\textbf{Model}} & \multicolumn{1}{c}{frame size $320\times 240$} & \multicolumn{1}{c}{frame size $640 \times 480$} & \multicolumn{1}{c}{\textbf{Subtask}} \\
\midrule
\textsc{SAM} (points\_per\_side = 16) & 0.71 & 0.58  {} &\\ %
\textsc{SAM 16bit} (points\_per\_side = 16) &  0.97  & 0.71 & Segmentation  \\
\textsc{FastSAM} & 10.7 & 10.2  &\\
\midrule
\textsc{DINO} & 4.76  & 4.68 &\\
\textsc{DINO Traced} & 6.27  & NA & Feature extraction\\
\textsc{DINO 16bit} & 10.63 & 11.55 & for click/image queries\\
\textsc{DINO 16bit+Traced} & 17.46 & 17.44 & \\
\midrule
\textsc{CLIP}  & 7.81 & 7.65 & Feature extraction\\
\textsc{CLIP 16bit}& 21.12& 20.21 & for text queries\\
\midrule
\textsc{SiamMask} & 50.3 & 49.2 & Tracking  \\
\textsc{DeAOT} & 28.74 & 17.13 & \\
\bottomrule
\end{tabular}}
\end{table}

\begin{table}[t]
\centering
\caption{Runtime in frames per second (FPS) for the detection phase of the system on an NVIDIA GeForce RTX 2070, compared to the popular open-source library~\cite{liu2023grounding}, using a more powerful GPU of NVIDIA RTX 3090.}\label{table:compete}
\adjustbox{max width=0.5\textwidth}
{
\begin{tabular}{lcc}
\toprule
\multicolumn{1}{c}{} & \multicolumn{1}{c}{\textbf{FPS}} & \multicolumn{1}{c}{\textbf{FPS}}  \\  
\multicolumn{1}{c}{\textbf{Approach}} & \multicolumn{1}{c}{frame size $320\times 240$} & \multicolumn{1}{c}{frame size $640 \times 480$}  \\ 
\midrule
\textsc{Ours with SAM} & 0.601 & 0.536  {} \\ %
\textsc{Ours with FastSAM} &  9.153  & 9.172   \\
\textsc{Ours with just DINO (no SAM or FastSAM)} & 15.67 & 14.37  \\
\midrule
\textsc{Grounded-SAM~\cite{liu2023grounding}} & 0.508 & 0.51  \\
\bottomrule
\end{tabular}}
\end{table}

%% file: text/03-experiments.tex
\section{Experiments}
\label{sec:exp}

We conducted several quadrotor experiments for zero-shot detection, tracking, and following different objects. We first outline key details of our system.
\subsection{Implementation and system details.} 

\noindent\textbf{Hardware.} We use a quadrotor equipped with an RGB camera (see Figure~\ref{fig:samcliplow}). The quadrotor is custom-built with a Pixhawk running PX4 flight control software. The camera data is streamed directly to a remote ground station computer equipped with an NVIDIA GeForce RTX 2070, and Intel i7-10750H CPU, with Ubuntu 20.04.5 LTS, using the ``herelink'' digital transmission system along with other telemetry data. 
The ground station runs the tracking algorithm and sends control commands to the quadrotor via Mavlink. To enable indoor testing, the quadrotor is also equipped with an onboard computer that runs MAVROS and interfaces with an external Vicon motion capture system to get the position. %

\begin{figure*}[t]
    \centering
    \subfigure[Drone following a drone]
    {
        \includegraphics[width=1\linewidth]%
        {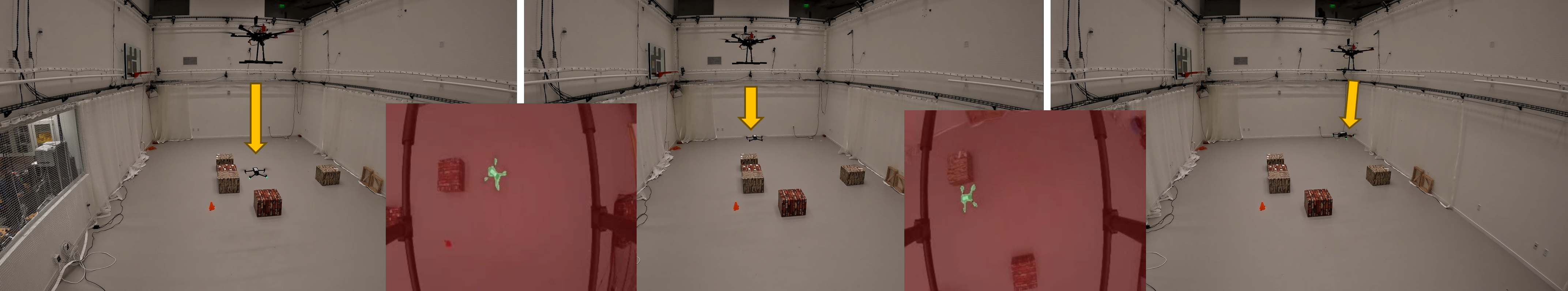}
        \label{fig:dronrefolowdrone}
    }
    \\
    
    \centering
    \subfigure[Drone following a toy car]
    {
        \includegraphics[width=1\linewidth]%
        {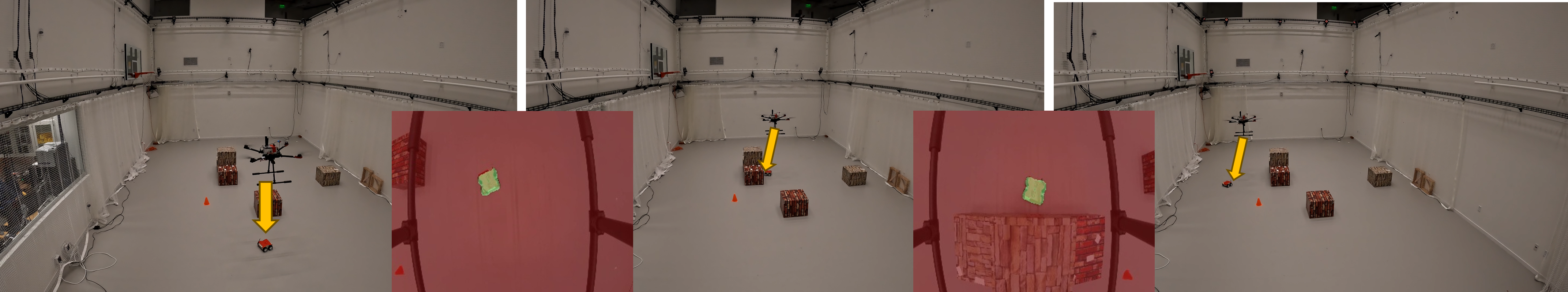}
        \label{fig:dronefollowcar}
    }
    \\ 
    \centering
     \subfigure[Drone following a toy (manually moved) brick]
    {
        \includegraphics[width=1\linewidth]%
        {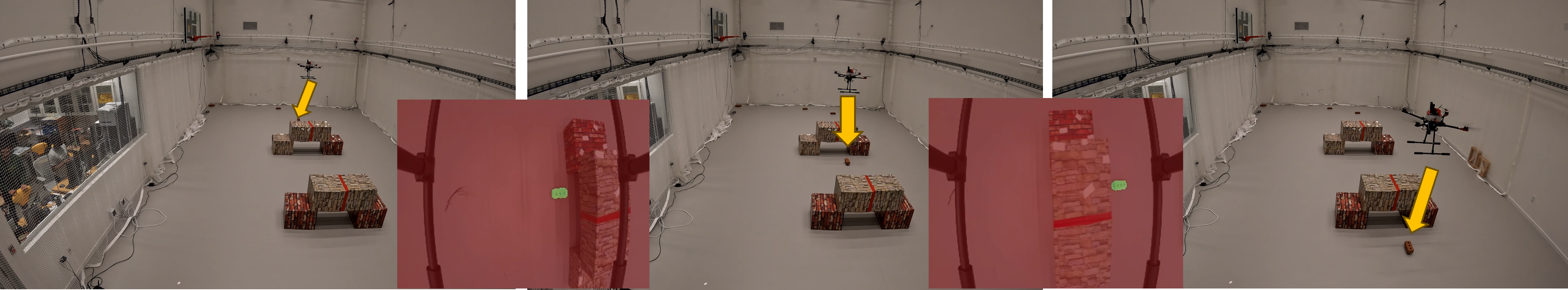}
        \label{fig:dronefollowbrick}
    }
     
     \caption{\textbf{Automatic tracking, following, and re-detection.} The tracked object is referred to by the yellow arrow, we also show the results of the re-detection mechanism in the last two rows.}
    \label{fig:sample_subfigures}
\end{figure*}

\noindent{{\textbf{Run-time improvement.}} We enhance segmentation/detection performance by compressing SAM and DINO through quantization and tracing and using FastSAM. For tracking, we offer support for the fast SiamMask~\cite{wang2019fast} tracker; see Table~\ref{table:FPS} for runtime (FPS) details.}

\noindent{\textbf{Flight controller.} For versatility, we used PX4, open-source flight control software, to interface with our quadrotor. The MAVSDK Python library is used to send velocity commands for 3D motion and yaw control, streamlining integration with PX4-based drones in future projects.} 

\noindent{\textbf{Visual servoing.} We mount the onboard camera on the bottom of the quadrotor facing the ground. At relatively small translational velocities the first-order approximations of roll and pitch angles are close to zero. In addition, we fixed the drone altitude and yaw angle. This simplifies the visual servoing task to 2D plane tracking using proportional control. We use a proportional controller based on pixel distances to center the object in the frame and employ a lowpass filter to smooth quadrotor trajectories, ensuring accuracy in challenging scenes.} 

\noindent{\textbf{Video Streaming. }
To process frames from an online video stream in real-time, we implemented a low-latency online streamer using the OpenCV library in Python. This streamer continuously reads frames with a parallel thread %
and maintains a buffer size of 1, ensuring immediate access to the latest frame when needed.} 

\noindent{\textbf{Software.} We mainly use Torch, cv2, and mavsdk; see our \href{https://github.com/alaamaalouf/FollowAnything}{project page} for full details. %

\subsection{Real time object following exprements} 

We tested (i) \new{our overall system for detecting, tracking, re-detecting, and following: RC cars, drones, and bricks in real-time. 
Here we used SAM+DINO and DINO-SOLO approaches for the detection task on all of the tested objects - the provided queries are clicks on the desired objects from other pictures (we provide a script for obtaining these click queries). Both approaches worked seamlessly for detecting and tracking the desired objects. } (ii)
We demonstrate our system \textbf{for re-detecting} an object that gets occluded from the scene during tracking. Specifically, during the following experiments, the RC car and the brick pass under a ``tunnel'' twice, and our re-detection mechanism is able to recover and resume tracking. Figures~\ref{fig:dronrefolowdrone},~\ref{fig:dronefollowcar}, and~\ref{fig:dronefollowbrick} show different scenarios during the following. We encourage the reader to view the demos on our \href{https://github.com/alaamaalouf/FollowAnything}{project webpage} and in the \href{https://www.youtube.com/watch?v=6Mgt3EPytrw}{explainer video}. (iii) \fix{In addition, we recorded the actual \textbf{3D trajectory} coordinates of the following quadrotor and the target object to assess the robustness of our tracking system. Specifically, we recorded continuous tracking data for over $4$ minutes while following a ground robot. We report the mean Euclidean distance between every point in the $x,y$ plane of the quadrotor and %
its aligned point in the plane (closest point) of the followed object. This experiment was conducted 4 times; using PID vs proportional-only as a controller, and using SAM+DINO vs DINO-SOLO as a detector. The results are reported in Figure~\ref{fig:control}. We can see that the drone follows the object smoothly and accurately using the different controllers and detectors. We also visualize both trajectories for the case of SAM+DINO.}

\begin{figure}[t]
    \centering
    \includegraphics[width=1\linewidth]{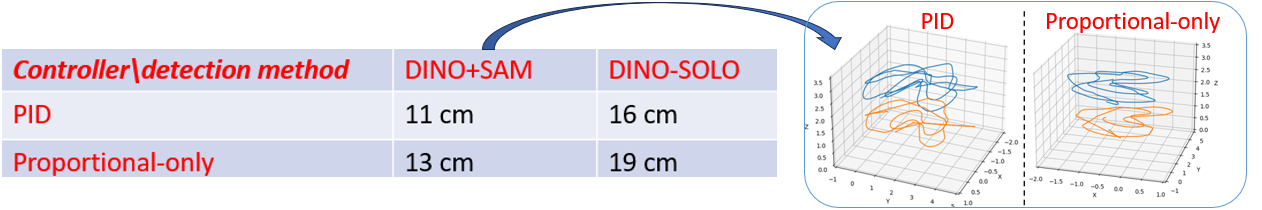}
    \caption{\fix{\textbf{Trajectory comparison.} We report the mean Euclidean distance between every point in the $x,y$ plane of the quadrotor and its aligned point in the plane (closest point) of the followed object.}} %
    \label{fig:control}
\end{figure}

\subsection{Zero-shot detection exprements}

\noindent\new{\textbf{Data.} We stored the tracking and detection streams from the SAM+DINO following experiment and used it to test the \FAM{} system and its different variants for zero-shot detection. For each of the tested objects, we picked multiple frames during the tracking and detection showcasing diverse object positions and diverse scenes. 
Other than that, we also use our private set of whale images to test on. }

\noindent\new{\textbf{Comparison.} We quantitatively compare the suggested methods and analyze their advantages and disadvantages. We applied each of SAM+DINO, SAM+CLIP, and DINO-SOLO to assess their efficacy in detecting the object within the given data. We report both True Positive and False Positive detection results. Furthermore, we conducted a comparative analysis involving an alternative version of our approach, which consists of two variations. (i) Majority Voting: We assigned each pixel in the mask to its most similar query, and subsequently assigned the mask the label that was most frequently selected across all mask pixels. (ii) $K$-Means: For each mask, we retained a set of $K>1$ representatives based on the $K$-means algorithm. We then gauged the similarity of these representatives with the provided queries and assigned the mask a label based on the majority consensus among these $K$ representatives. 
Our testing encompassed two scenarios: (i) The system was presented with multiple queries representing the environment, including "a robot leg, a box, a ground" (in the whales experiment, these queries were replaced with "water"), along with the desired query "a drone, a toy car, a brick, a whale" (Table~\ref{table:compete_multiple_queries}) and (ii) The system was given a single desired query (Table~\ref{table:compete_single_query}).}  

\noindent\new{\textbf{The threshold $\alpha$.} For all methods, we tuned $\alpha$ to minimize the false positive detections while achieving a fine true positive detection rate; In our system, it's acceptable to not immediately identify the intended object, but our priority is to prevent the detection and tracking of an incorrect target. 
We used $0.35$ for SAM+DINO and $0.23$ for DINO+CLP in all experiments. For DINO-SOLO $0.4$ and $0.6$ were used in the multiple queries and single query experiments, respectively.}

\fix{
\subsection{Mask quality experements}

We compare the mask quality of our detection methods (DINO-SOLO, SAM+DINO/CLIP). We use the first video from the Cholec80 dataset~\cite{hong2020cholecseg8k}, which has mask annotation for body parts and tools across frames during surgery. We aimed to detect the “grasper” tool and track it across frames. Table~\ref{table:TRACKMAS} reports (1) the mean intersection over union (mIoU) of the detection and the annotated data across frames and (2) the true positive detection percentage of the desired object, we also test how the detected mask quality affects the tracking; we report (3) the mIoU of the desired tracked object after each of the detection methods detected it. Queries: "body part", "background", and "surgery tool".}%

\begin{table}[t]
\centering
\caption{\fix{true positive detections divided by the number of object appearances (provided next to the object name), and number of false positive detections (number in brackets if any); single query test.}}\label{table:compete_single_query}
\adjustbox{max width=0.5\textwidth}
{
\begin{tabular}{lcccc}
\toprule
\textbf{Approach} & \textbf{Car} (11) & \textbf{Drone} (15) & \textbf{Bricks} (15)& \textbf{Whales} (25)   \\
\midrule
\textsc{SAM+DINO} &  1.0     & 0.5   &    0.6    & 0.84 (2)     \\ %
\textsc{SAM+CLIP} & 0.81 (3)  & 0.4 (1)  &  NA &  0.8 (1)   \\
\textsc{DINO-SOLO} &  0.91 & 0.7    &   0.73  & 0.92 (1)     \\
\textsc{10-MEANS} & 1.0   	&  0.5    &  0.6   &   0.84 (3)    \\
\textsc{5-MEANS} &   0.91   & 0.4  &     0.53   &   0.8 (2)   \\
\textsc{MAJORITY  VOTING} & 1.0   & 0.5     &    0.53    &   0.84 (3)   \\
\bottomrule
\end{tabular}}
\end{table}

\subsection{Discussion and conclusions}

\noindent\textbf{SAM+DINO.} Figures~\ref{fig:autodetect} and~\ref{fig:example} show example results for real-time detection via SAM+DINO. Tables~\ref{table:compete_multiple_queries} and~\ref{table:compete_single_query}, indicate that the detection achieves a high level of accuracy for cars and whales, and performs well for drones and bricks - but may occasionally miss certain instances. \new{After analyzing the results, it becomes apparent that when SAM generates reliable regions/segmentation, DINO consistently assigns the correct labels to each of these regions, ensuring precise and appropriate object detection. However, in cases where SAM fails to capture these regions accurately (resulting in inadequate segmentations), the object goes undetected.  This scenario is exemplified by $4$ drone object in the dataset and $3$ bricks, where SAM fails to identify the mask of the drone/brick (see Figure~\ref{fig:sampvsnosam} for example). Regarding accurate DINO classifications, we offer explanations illustrated in Figure~\ref{fig:heatmaps}. These figures depict heatmaps based on cosine similarity calculations between DINO feature descriptors of each pixel and a designated focal point pixel. The visualizations clearly demonstrate that pixels sharing similar semantic characteristics exhibit a high degree of similarity in their DINO features. 
}

\begin{table}[t]
\centering
\caption{\fix{true positive detections divided by the number of object appearances (provided next to the object name), and number of false positive detections (number in brackets if any); multiple queries test.}}\label{table:compete_multiple_queries}
\adjustbox{max width=0.5\textwidth}
{
\begin{tabular}{lcccc}
\toprule
\textbf{Approach} & \textbf{Car} (11) & \textbf{Drone} (10) & \textbf{Bricks} (15) & \textbf{Whales} (25)   \\
\midrule
\textsc{SAM+DINO} &  1.0 & 0.6  &  0.67  &  0.84   \\ %
\textsc{SAM+CLIP} &  0.91 (2) & 0.5 (1)  &   0.4 (5)& 0.65    \\
\textsc{DINO-SOLO} &   1.0  & 0.9   &  0.87 & 0.92 (1)    \\
\textsc{10-MEANS} &    1.0	& 0.6  &   0.67 &   0.8   \\
\textsc{5-MEANS} & 1.0  &  0.5  &     0.6   &  0.8    \\
\textsc{MAJORITY  VOTING} &  1.0 &   0.6  & 0.67    & 0.84     \\
\bottomrule
\end{tabular}}
\end{table}

\begin{figure}[t]
    \centering
    \includegraphics[width=1\linewidth]{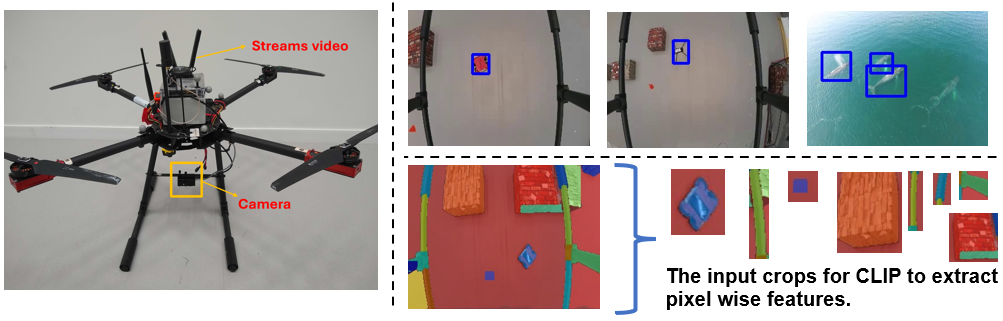}
     \caption{\textbf{Left:} Our custom-built quadrotor.\new{ \textbf{ Right-up:} Successful automatic detection via text queries (SAM+CLIP) on low-resolution images; text queries used from left to right: "a toy car" (single query), "a drone" (single query), and "a whale"+"water" (multiclass). \textbf{Right bottom:} In some cases, the raw data from the cropped masks (to get pixel-wise features from CLIP) does not provide enough information for CLIP - since the image is of low resolution and the mask is small causing it to provide not accurate descriptors and thus \FAM{} may not detect the objects.}}
    \label{fig:samcliplow}
\end{figure}

\begin{table}[t]
\centering
\caption{\fix{mIoU for tracking and detection (DINO-SOLO vs SAM+DINO or CLIP). We also report the accuracy of the detection (percentage of true positive detections from all of the parsed frames). Note that in the tracking SAM+DINO and SAM+CLIP got the same results as they were provided the same mask by SAM (the first detected). The used tracker is DEAOT.}}\label{table:TRACKMAS}
\adjustbox{max width=0.5\textwidth}
{
\begin{tabular}{lcccccc}
\toprule
\textbf{Approach}  & \textbf{Stage} & \textbf{mIoU (480x854)} & \textbf{mIoU (240x427)} & \textbf{Acc (480x854)}   & \textbf{Acc (240x427)}\\
\midrule
\textsc{SAM+DINO}    &  Tracking &    0.87   & 0.77  &NA&NA   \\ %
\textsc{SAM+CLIP}    &  Tracking &    0.87   & 0.77   &NA&NA \\ %
\textsc{DINO-SOLO}    &  Tracking &   0.84   & 0.75     &NA&NA\\ %
\midrule
\textsc{SAM+DINO}    &  Detection &    0.86    & 0.81     &0.89&0.89\\ %
\textsc{SAM+CLIP}    &  Detection &    0.82    & 0.68   &0.18&0.13  \\ %
\textsc{DINO-SOLO}    &  Detection &    0.73    & 0.67     &0.98 &0.98\\ %
\bottomrule
\end{tabular}}
\end{table}
\begin{figure}[t]
    \centering
    \includegraphics[width=1\linewidth]{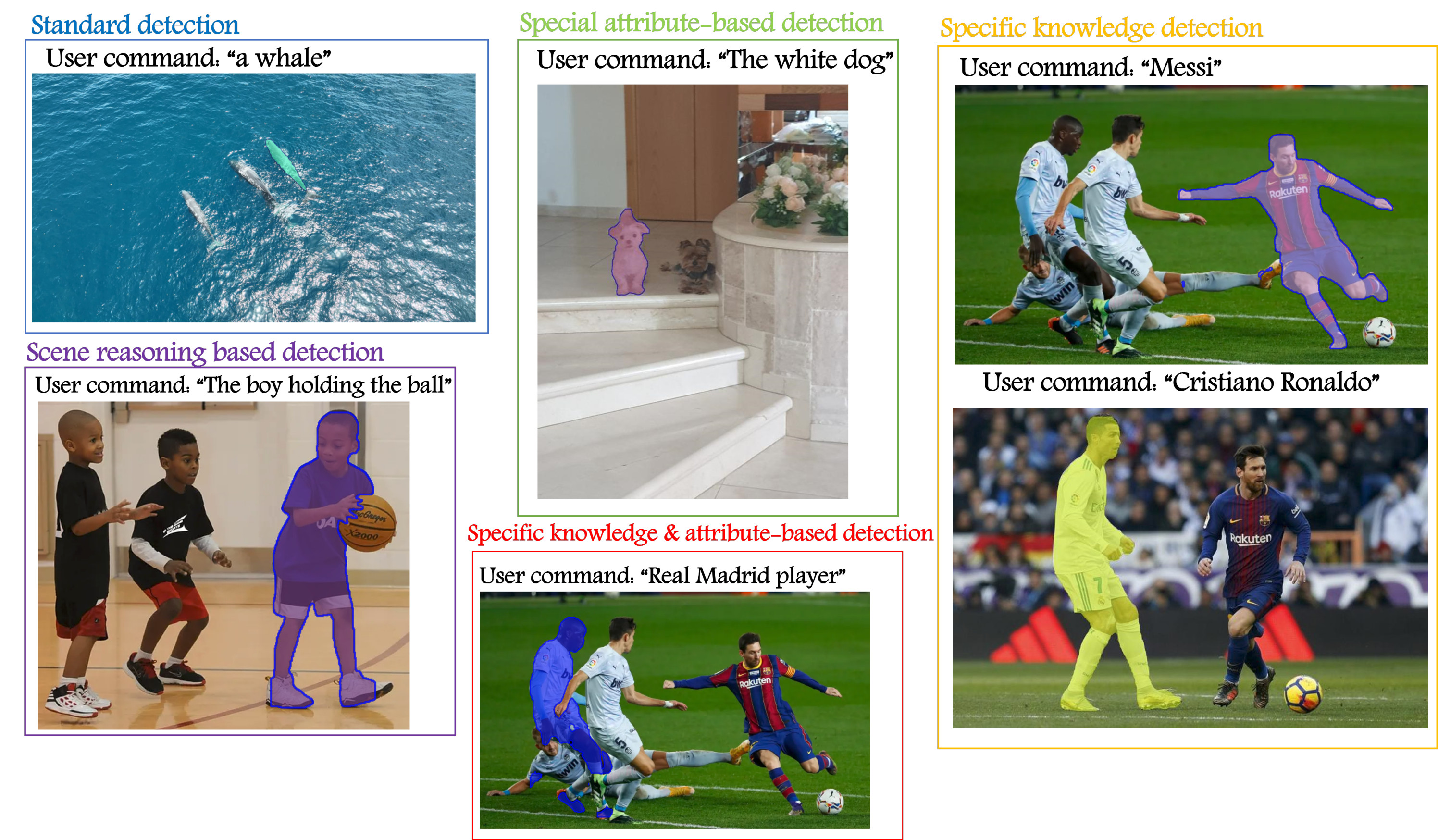}
    
\caption{{Automatic detection experiments via text queries (SAM-and-Clip) on high resolution data.}} %
    \label{fig:clipdetect}
\end{figure}

\begin{figure}[t]
    \centering
    \includegraphics[width=\linewidth]{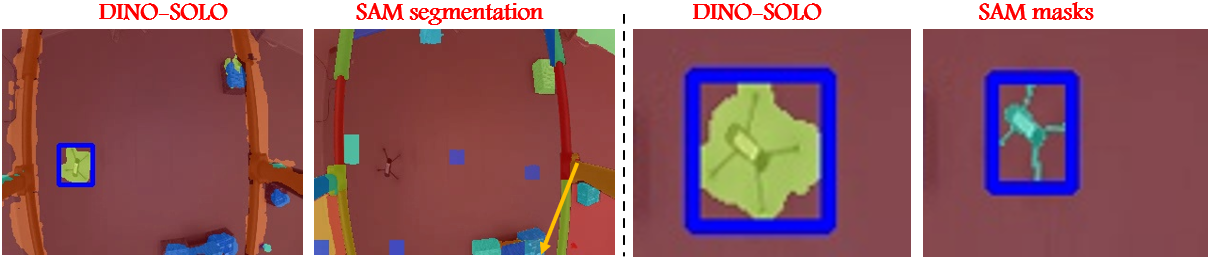}
    \caption{\new{\textbf{With vs without SAM.} Right: SAM creates high-quality segmentation masks compared to DINO-SOLO (not using SAM). Left: SAM might miss important regions in the image.}} 
    \label{fig:sampvsnosam}
\end{figure}

\noindent\textbf{DINO-SOLO.} In Figure~\ref{fig:fastapp} we show several examples showcasing the efficiency of our rapid automated detection system. 
\new{This approach is significantly faster and performs admirably in detecting the desired objects. Even more, in many cases when SAM misses providing the desired object a mask, using DINO-SOLO can still detect the object. However, the resulting masks are not of high quality compared to the masks obtained from SAM, and this may potentially affect the tracking performance; see Figure~\ref{fig:sampvsnosam} and Table~\ref{table:TRACKMAS}.}

\noindent\new{\textbf{SAM+CLIP\footnote{The single query CLIP brick detection experiment was not conducted due to using a single query in an environment with similar objects.}. } Examples for detection via "text" prompts through SAM+CLIP are shown in Figure~\ref{fig:samcliplow}. For the tested low-resolution images, SAM+CLIP detections are not as robust, the method yields less precise similarity scores, increasing the likelihood of missed detections, particularly for objects lacking unique shapes like the brick. Additionally, in some cases, as the image has low resolution, if the object has a small (correct) mask, it does not present enough raw information and is thus misclassified. Figure~\ref{fig:samcliplow} shows an example of low-resolution images for such scenarios; we further discuss why this happens when using CLIP and not DINO in our discussion later. We note that this method is still beneficial for our system, the main idea is that we need one accurate detection with high confidence (e.g., with further increasing $\alpha$) for the desired object and then we can start the object-following scheme, thus, we can still benefit from the multimodality of the system. Additionally, as this method requires only the text prompt and not an image/clicks, it is much easier to utilize.} %
\new{To verify our claims regarding the reason for the dropped performance of \FAM{} when using SAM+CLIP, we tested it on high-resolution images. Here, the reasoning and detection are robust justifying our claims.} We conduct the following 4 tests: 
(i) {Standard detection}, e.g., ``detect a whale'', (ii) {scene reasoning-based detection,} %
e.g., ``detect the boy \emph{holding the ball}''. (ii) {Special attribute-based detection,} %
like, ``detect the \emph{white} dog''. (iv) {Special prior knowledge-based detection.} In this case, the system should have prior knowledge of a specific object like its name/nickname. For example ``detect Messi/Cristiano Ronaldo''.  (v) {Special prior knowledge\& attribute based detection.} %
e.g., ``detect a Real Madrid player''.
See result in Figure~\ref{fig:clipdetect}.

\noindent\new{\textbf{SAM limitations: With vs without.} SAM might miss important regions in the image. When the desired object is in these regions it will be impossible to detect it and thus DINO+SAM yields fewer true positive detections compared to  DINO-SOLO. On the other hand, DINO+SAM provides high-quality masks once the object is detected while DINO-SOLO masks are less refined. \fix{See Tables~\ref{table:TRACKMAS},~\ref{table:compete_single_query}, and~\ref{table:compete_multiple_queries}.}}

\noindent\new{\textbf{Queries.} Using \textbf{multiple} queries to annotate other objects that might be in the scene reduces the number of False positives leading to a more robust and reliable system. }

\noindent\new{\textbf{DINO vs CLIP}. The method we are using to obtain pixel-wise features from DINO~\cite{amir2021deep} is faster and provides better descriptors for every pixel compared to the method used for CLIP. This is because it requires one forward pass on the whole image to compute the per-pixel features. In addition, when using DINO, the method computes the per patch/pixel features while taking into count the full image, as it simply utilizes the patch-wise descriptors (outputs of the query, key, or value matrix in some attention layer of the transformer) of DINO, thus providing descriptors with richer context of the whole image. In CLIP, the method uses SAM to extract masks~\cite{jatavallabhula2023conceptfusion} and then applies CLIP on crops of these masks to extract features for all pixels in this mask, thus, it is less efficient and might yield less meaningful features when applying CLIP on small crops with limited raw data.}%

\noindent\new{\textbf{The competing methods.} 
We found no improvement with other variants like $K$-means and majority voting; often, our original methods performed better. \fix{Also, the $K$-means variant runs at 0.03 FPS, and the majority voting runs at 0.32 FPS.}}

%% file: text/04-discussion-and-conclusion.tex
\noindent\textbf{{Summary.}} 
 \FAM{} bridges the gap between SOTA computer vision and robotic systems, providing an end-to-end solution for detecting, tracking, and following any object. Its open set, multimodal, real-time capabilities, adaptability to different environments, and open-source code make it a valuable tool.

\section*{Acknowledgment}
This study was funded by Project CETI via grants from Dalio Philanthropies and Ocean X; Sea Grape Foundation; Virgin Unite, Rosamund Zander/Hansjorg Wyss, Chris Anderson/Jacqueline Novogratz through The Audacious Project: a collaborative funding initiative housed at TED. 
This research was supported in part by the AI2050 program at Schmidt Futures (grant G-22-63172).